\documentclass{article}

\usepackage{arxiv}

\usepackage[utf8]{inputenc} % allow utf-8 input
\usepackage[T1]{fontenc}    % use 8-bit T1 fonts
\usepackage{hyperref}       % hyperlinks
\usepackage{url}            % simple URL typesetting
\usepackage{booktabs}       % professional-quality tables
\usepackage[table]{xcolor} % color for tables
\usepackage{amsfonts}       % blackboard math symbols
\usepackage{nicefrac}       % compact symbols for 1/2, etc.
\usepackage{microtype}      % microtypography
\usepackage{lipsum}		% Can be removed after putting your text content
\usepackage{graphicx}
\usepackage{natbib}
\usepackage{doi}
\usepackage{multirow}
\usepackage{makecell}

\title{Generalization of Text Line Segmentation for HTR in Historical Documents}

\author{ \href{https://orcid.org/0009-0000-9164-487X}{\includegraphics[scale=0.06]{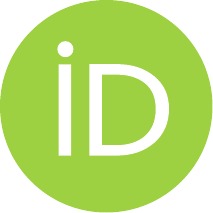}\hspace{1mm}Gayan H. Pathirage} \\
	Machine Learning Group\\
    Luleå University of Technology\\
	Luleå, Sweden\\
	\texttt{gayan.pathirage@ltu.se} \\
	\And
	\href{https://orcid.org/0000-0003-4796-7182}{\includegraphics[scale=0.06]{orcid.pdf}\hspace{1mm}Stephan M. Unter} \\
    Machine Learning Group\\
	Luleå University of Technology\\
	Luleå, Sweden\\
	\texttt{stephan.unter@associated.ltu.se} \\
	\And
	\href{https://orcid.org/0000-0002-8788-7198}{\includegraphics[scale=0.06]{orcid.pdf}\hspace{1mm}Simon Corbillé} \\
	University of Rouen Normandy\\
	Rouen, France\\
	\texttt{simon.corbille@univ-rouen.fr} \\
	\And
	\href{https://orcid.org/0000-0003-2039-3844}{\includegraphics[scale=0.06]{orcid.pdf}\hspace{1mm}Elisa~H.~Barney~Smith} \\
    Machine Learning Group\\
	Luleå University of Technology\\
	Luleå, Sweden\\
	\texttt{elisa.barney@ltu.se} \\
}

\renewcommand{\headeright}{preprint}
\renewcommand{\shorttitle}{\textit{Generalization of Text Line Segmentation for HTR in Historical Documents}}

\hypersetup{
pdftitle={Generalization of Text Line Segmentation for HTR in Historical Documents},
pdfauthor={Gayan H. Pathirage, Stephan M. Unter, Simon Corbillé, Elisa H. Barney Smith},
pdfkeywords={Text Line Segmentation, Handwritten Text Recognition (HTR), Domain Generalization, Historical Document Analysis},
}

\begin{document}
\maketitle

\begin{abstract}
	In this paper we investigate the generalization capability of state-of-the-art object detection models (Hi-SAM, YOLO, and RT-DETR) for the automatic segmentation of text lines in historical documents. 
Generalization is needed because it is not practical for museums and archives to produce large amounts of labeled samples for each new collection.
We also evaluate the few-shot setting by fine-tuning the networks with only a small number of training samples per target dataset. 
To this end, we use nine different datasets that differ significantly in visual layout and language.
Additionally, we discuss why mean average precision (mAP), a classic evaluation metric for object detection, is insufficient for assessing the quality of a text line segmentation model. 
Depending on the dataset, our CRNN models achieve character error rates (CER) ranging from 2.48\% to 6.50\% per page for the segmented text lines. This is sufficient for reading comprehension and simple text searches. Finally, we quantify the resource and time requirements of training and inference in different use case settings. This forms the basis for more informed decisions about model and hardware choices based on available resources.
\end{abstract}

% keywords can be removed

\section{Introduction}
\label{sec:introduction}

The automated recognition of digitized handwritten documents, particularly historical texts, remains a central research question and challenge in document analysis. One particular difficulty is not only the general state of preservation of historical documents, which can make them difficult to read for both humans and machines, but also the visual differences in handwriting, language, appearance and layout. These differ from modern versions in semantics and character sets and pose significant challenges for machine-based systems.

Although recent research has focused on developing systems that combine layout analysis and text recognition for whole document pages \cite{coquenet2023dan}, a multi-step process remains a dominant approach. This begins with the localization and segmentation of text lines in the page, after which the contents of each line are converted into digital text via a handwritten text recognition (HTR) model.

There are several page level datasets researchers use for experiments. These generally have many pages of carefully labeled text zones and proofed textual content. These can be used to train or retrain a system to achieve a high quality recognition. When a library or archive is interested in digitizing and recognizing its contents, it wants to do so with as little careful zoning and transcription as possible. A set of tools is needed that has good performance as is, or that can be adapted with very few labeled pages.

This raises the question of generalizability: Is it necessary to train new models for each new dataset? Could the information learned from established datasets be easily transferred to new ones? To what extent can trained "foundation" models generalize to unseen data in contexts that are comparable yet not identical to historical datasets?

To the best of our knowledge, no in-depth analysis of this issue for historical documents has yet been conducted. Hence, in our paper we aim to more closely analyze whether and to what extent text line detection models built from existing, well-established source datasets work for new target domains that differ significantly in terms of visual appearance and content. To this end, we evaluate three line detection methods: Hi-SAM -- a model developed specifically for identifying elements in scene-text datasets containing printed and handwritten text in natural-image environments; and YOLO and RT-DETR -- methods that have generally delivered robust results for detecting elements in images. We consider their direct generalizability to new document datasets, and then evaluate how retraining with a small portion of a new dataset affects their performance. We further explore the usefulness of the mean average precision (mAP) evaluation metric, typically used with detection networks, for assessing the quality of text line segmentation networks with the downstream task of HTR in mind.

Modern line segmentation models can be large and computationally expensive. This is another challenge for institutions like archives and libraries who usually will not have access to large supercomputing networks. 
Therefore, we further report the memory and compute needed to train and do inference with these models and provide an overview of the necessary resources, including RAM, training time, and inference time, for different use cases, ranging from a regular laptop without a dedicated GPU to a high-performance cluster designed for machine learning tasks.

Hence, the contributions of our paper are as follows:

\begin{itemize}
    \item a study on the generalization capability of established object detectors for the purpose of text line segmentation, evaluated on multiple varying target datasets,
    \item a detailed view into standard evaluation metrics for text line segmentation with respect to the downstream end goal of text recognition,
    \item an analysis of the impact of finetuning pretrained networks on small numbers of pages per target domain,
    \item an overview of computational and data requirements for producing competitive results in low-resource scenarios.
\end{itemize}

In Section \ref{sec:related-work} we overview research in the fields of text line segmentation. Section \ref{sec:experiments} describes the datasets used in this work, the segmentation and recognition models and metrics as well as the experimental setup. The corresponding results are detailed in Section~\ref{sec:results}. An in-depth discussion on the results is conducted in Section \ref{sec:discussion}, with conclusions in Section~\ref{sec:conclusion}.

\section{Related Work}
\label{sec:related-work}

Numerous methods have been proposed to detect and delineate text lines in document images. Traditional approaches often rely on structural assumptions, such as text being predominantly horizontal and straight, and the presence of visible gaps between adjacent lines where the lighter background separates darker ink regions. 
Classical line segmentation techniques include adaptive partial projection, matched filtering, midpoint detection, thinning-based methods, second-order derivative analysis, and path-finding approaches. These techniques generally rely on handcrafted heuristics and structural assumptions regarding text alignment, baseline regularity, and inter-line spacing and require careful parameter tuning to handle complex scripts and degraded manuscripts~\cite{Likforman07,ratheash2019detailed}.

Although such classical techniques remain computationally efficient and interpretable, their robustness is often limited when confronted with irregular baselines, variable ascender and descender lengths, and characters overlapping across neighboring lines. Historical documents further introduce challenges such as background degradation, yellowing, and bleed-through artifacts, which complicate reliable line separation. As a result, the robustness of rule-based techniques is limited when confronted with such variability.

To address these limitations, recent research predominantly uses machine learning-based methods which are data-driven and rely on learned representations from annotated data rather than handcrafted structural assumptions~\cite{Boillet22,Rabaev26}.
These are therefore better suited to handle diverse document characteristics. In this work, we investigate several modern architectures, including general-purpose instance segmentation models, object detection frameworks, and a model specifically designed for modern text-line segmentation.

Hi-SAM~\cite{ye2024hi}, a transformer-based architecture designed for word, text-line, and paragraph-level segmentation. The model builds upon the Segment Anything Model (SAM)~\cite{kirillov2023segment}, which typically requires user-provided prompts to guide segmentation. To eliminate this dependency, Hi-SAM incorporates a self-prompting mechanism that automatically generates prompts for relevant regions of interest.

You Only Look Once (YOLO)\footnote{\url{https://docs.ultralytics.com/models/}, last accessed 26 May 2026.} is a real-time object detection framework that can be adapted for text line localization by predicting bounding boxes or segmentation masks. Versions, starting with YOLOv8, which support instance segmentation, enable pixel-level delineation of text regions instead of relying on rectangular bounding boxes. This is particularly beneficial for overlapping lines, non-horizontal layouts, and curved text-lines.

RT-DETR~\cite{RT-DETR} is an end-to-end, transformer-based real-time object detection framework that removes the reliance on Non-Maximum Suppression (NMS) enabling it to achieve superior performance compared to YOLO-based detectors in real-time settings. The model architecture comprises an efficient hybrid encoder coupled with a transformer decoder enhanced by auxiliary prediction heads.

Despite their strong performance on benchmark datasets, the cross-dataset generalization of these architectures for handwritten document analysis remains insufficiently studied. In particular, the \textbf{relationship between detection metrics and downstream recognition performance} has not been systematically examined. In this work, we address this gap by evaluating YOLOv8, Hi-SAM, and RT-DETR under domain shifts and assess segmentation quality using both detection and recognition-based metrics.

\section{Experimental Protocol}
\label{sec:experiments}

\subsection{Datasets}
\label{sec:datasets}

\begin{figure}[!htbp]
    \centering
    \includegraphics[width=1\linewidth]{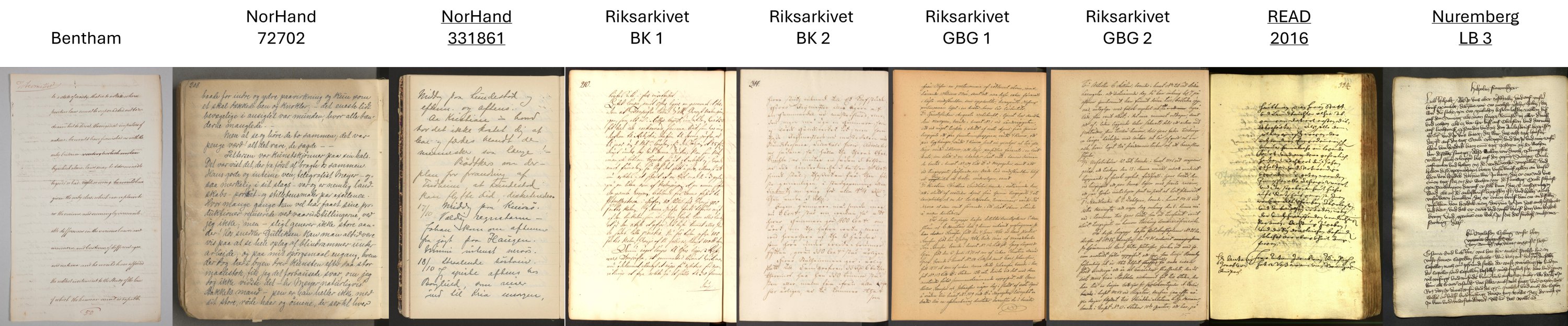}
    \caption{Visual overview of all used datasets. The documents vary in language, handwriting style, and paper color. Some documents are more prone to bleed-through effects than others.}
    \label{fig:overview_datasets}
\end{figure}

Our study is based on nine historical datasets with diverse visual and linguistic characteristics. Table~\ref{tab:dataset_statistics} summarizes the dataset statistics, while Figure~\ref{fig:overview_datasets} provides a visual overview. Three datasets are selected as source domains for training based on their visual diversity and comparable number of pages to ensure a fair comparison.

\begin{table}[!htbp]
\centering
\caption{Dataset statistics for the ten datasets used in the study. The table reports the number of pages and text lines in the training, validation, test, and total splits. Underlined datasets have been used as source domains for training.}
\label{tab:dataset_statistics}
\resizebox{\textwidth}{!}{%
\begin{tabular}{l|l|c|c|c|c|c|c|c|c|c|c}
\multicolumn{2}{c|}{}
& \multicolumn{1}{c|}{\cellcolor{gray!20}\textbf{English}}
& \multicolumn{2}{c|}{\cellcolor{gray!20}\textbf{Norwegian}}
& \multicolumn{4}{c|}{\cellcolor{gray!20}\textbf{Swedish}}
& \multicolumn{2}{c|}{\cellcolor{gray!20}\textbf{German}}
& \multicolumn{1}{c}{\cellcolor{gray!20}\textbf{Mix Dataset}} \\
\hline
\multicolumn{2}{c|}{}
& \makecell{\\\textbf{Bentham}\\\cite{sanchez_bentham_2016}}
& \makecell{\textbf{NorHand}\\\textbf{72702}\\\cite{NorHandv3}}
& \makecell{\underline{\textbf{NorHand}}\\\underline{\textbf{331861}}\\\cite{NorHandv3}}
& \makecell{\textbf{Riksarkivet}\\\textbf{BK 1}\\\cite{Riksarkivet}}
& \makecell{\textbf{Riksarkivet}\\\textbf{BK 2}\\\cite{Riksarkivet}}
& \makecell{\textbf{Riksarkivet}\\\textbf{GBG 1}\\\cite{Riksarkivet}}
& \makecell{\textbf{Riksarkivet}\\\textbf{GBG 2}\\\cite{Riksarkivet}}
& \makecell{\underline{\textbf{READ}}\\\underline{\textbf{2016}}\\\cite{sanchez2016icfhr2016}}
& \makecell{\underline{\textbf{Nuremberg}}\\\underline{\textbf{LB 3}}\\\cite{mayr_nuremberg_2025}}
& \underline{\textbf{Mix}} \\
\hline
\multicolumn{2}{l|}{Time} & 19th century& \multicolumn{2}{c|}{19th/20th century}& \multicolumn{2}{c|}{18th century}& \multicolumn{2}{c|}{19th century}& 15th-19th century& 15th century&\\
\hline
\multirow{4}{*}{\rotatebox{90}{\textbf{Pages}}}
& Train & 350 & 292 & 319 & 184 & 175 & 208 & 146 & 350 & 433 & 336 \\
& Val   & 50  & 37  & 40  & 23  & 22  & 26  & 18  & 50  & 55  & 47  \\
& Test  & 33  & 37  & 40  & 24  & 22  & 27  & 19  & 50  & 53  & 46  \\
& Total & 433 & 366 & 399 & 231 & 219 & 261 & 183 & 450 & 541 & 429 \\
\hline
\multirow{4}{*}{\rotatebox{90}{\textbf{Lines}}}
& Train & 9,198  & 6,069 & 6,053 & 8,824  & 9,510  & 14,166 & 10,954 & 8,367  & 12,820 & 9,067  \\
& Val   & 1,415  & 792  & 784  & 1,147  & 1,201  & 1,753  & 1,319  & 1,043  & 1,759  & 1,196  \\
& Test  & 860   & 753  & 766  & 1,117  & 1,186  & 1,915  & 1,436  & 1,140  & 1,604  & 1,150  \\
& Total & 11,473 & 7,614 & 7,603 & 11,088 & 11,897 & 17,834 & 13,709 & 10,550 & 16,183 & 11,413 \\
\hline
\end{tabular}%
}
\end{table}

The first training dataset is part of an extensive collection of Norwegian manuscripts, \textbf{NorHand v3} \cite{NorHandv3}. It includes letters and diaries from the 19th and early 20th centuries. Most pages exhibit yellowish or brownish discoloration. Bleed-through occurs but is not problematic here. In contrast, the handwriting shows significant variance. Within the original dataset, individual books can be identified by numerical prefixes. We use book 331861 as a training dataset.

\textbf{READ 2016} \cite{sanchez2016icfhr2016} is a subset of pages collected from Ratsprotokolle's council minutes from the 15th to the 19th century. The documents also exhibit slight yellowish discoloration. The handwriting features sweeping, elongated letters, particularly at the start of paragraphs and in headings. Bleed-through, or text showing through from the other side of the page, poses an additional challenge.

The \textbf{Nuremberg LB 3} is drawn from a dataset of German-language documents dating from the early fifteenth century. Although we limited ourselves to the third volume to closely match the number of pages to the READ 2016 dataset, the original dataset \cite{mayr_nuremberg_2025} comprises a total of four books. The pages of these documents are less yellowed than those in READ. Most striking, however, is the highly ornate, rounded style of the handwriting. 

We hypothesized that a more diverse dataset might be better suited for training a foundation model. For this reason, we introduced an additional \textbf{Mix} dataset consisting of one-third each of the three previously described training datasets.

We selected test splits of six other books for generalizability testing and training and validation splits for fine tuning. 
In addition to \textbf{NorHand v3} book 331861 which is used as a training dataset, we use book 72702 for additional inference. 
We used four Swedish books from the Riksarkivet~\cite{Riksarkivet}:\footnote{To our knowledge, these texts were digitized automatically using an HTR model. Since the quality is sufficient for reading comprehension and searching, the texts have not been corrected in detail and may thus contain errors.} 
\textbf{BK~1} ("\textit{Bergskollegium Relationer och skrivelser E3 10 1718-1727}"),
\textbf{BK~2} ("\textit{Bergskollegium Relationer och skrivelser E3 11 1718-1727}"),
\textbf{GBG~1} ("\textit{Göteborgs poliskammare före 1900 A II 12 1880}"), and
\textbf{GBG~2} ("\textit{Göteborgs poliskammare före 1900 A II 1b 1869}").
In these four cases, the datasets consist of images showing two pages of a book. Since this study focuses exclusively on single-page document images, we used a simple algorithmic approach to split all pages in the center between given text line regions. We only had to adjust a couple of samples manually. We also used the established \textbf{Bentham} \cite{sanchez_bentham_2016} dataset as an example for an English manuscript. We  publish the new dataset splits\footnote{The splits for NorHand-331861, NorHand-72702, Nuremberg LB 3, Riksarkivet BK~1 and 2, and Riksarkivet GBG~1 and 2 are published on \url{https://doi.org/10.5281/zenodo.20289977}.}.

\subsection{Evaluation Metrics}
\label{sec: Evaluation Metrics}

We use \textbf{mean average precision} (mAP) as a metric to evaluate the performance of our object detection networks. It is defined as
\[
mAP = \frac{1}{n} \sum _{i=1}^{n} AP_i
\]
where $AP_i$ is the average precision for class $i$ (in our case, $n=1$ as we are only predicting text lines). The average precision is defined as the area under the precision-recall-curve. To define the alignment of predictions and ground-truth (GT) annotations, an Intersection over Union (IoU) threshold is determined. We compute the $mAP_{50:95}$ metric, which takes the average for all IoU thresholds from 0.50 to 0.95 in incremental steps of step width 0.05. The higher this value, the closer the alignment of predictions and GT boxes.

While mAP is commonly used to evaluate object detection models, we ultimately have text recognition in mind as the downstream task. Therefore, the quality of the recognition results is equally important, as we need to ensure that the predicted text line boxes provide suitable input for the HTR system to extract the text effectively.

We evaluate the quality of HTR with the Character Error Rate (CER):
\begin{equation}
    CER = \frac{S + I + D}{N} ,
    \label{eq:CER}
\end{equation}

\noindent which is based on the Levenshtein edit distance~\cite{yujian2007normalized}. CER quantifies the number of character-level operations required to transform the predicted text into the ground-truth. It is computed as the ratio of the total number of  substitutions (S), insertions (I), and deletions (D) edits to the number of characters (N) in the ground-truth. Lower CER values indicate better recognition performance, reflecting fewer discrepancies between the predicted and reference text.

To compute the CER for the predicted text lines, the corresponding ground-truth text must first be identified. Since the predicted bounding boxes are not guaranteed to follow the top-to-bottom reading order and some lines are not detected, we match the predicted boxes to the ground-truth boxes by first computing the IoU between them and then applying the Hungarian search algorithm to obtain the optimal assignment. This matching enables accurate line-level CER computation. The edit distance is computed for each matched line pair. The standard CER penalties for insertions and deletions are also applied when the number of predicted boxes differs from the number of ground-truth boxes, accounting for extra or missing detections. Finally, the edit distances are accumulated over all lines and pages in the split and then normalized by the total number of characters \(N\).

\subsection{Text Line Detection Models}
\label{sec:Line Detector Model Architectures}

\textbf{Hi-SAM}. We used the Hi-SAM model introduced in~\cite{ye2024hi} and adopted the official PyTorch implementation. The authors released three model variants (base, large, and huge). Based on our preliminary experiments, the huge variant (H) provided the best performance for text-line segmentation, therefore, we used it in this work. The model has five modules: an image encoder, a pixel-level mask decoder, and frozen prompt encoder derived from SAM~\cite{kirillov2023segment}, a self prompt module, and a hierarchical mask decoder. They included trainable adapters to the two Vision Transformer (ViT) blocks in the encoder but kept the encoder frozen.

\textbf{YOLO}. You Only Look Once (YOLO) is a well-established family of one-shot object detection models, nowadays also used for detecting and localizing entities in handwritten documents, such as individual characters. New models with incremental version numbers are constantly being released, but they do not necessarily represent direct improvements in HTR. Rather, they demonstrate various approaches and model architectures within the one-shot YOLO framework. Additionally, newer YOLO architectures have advanced capabilities, such as instance segmentation and oriented box detection. After testing various versions and variants in a preliminary experimental study, this paper focuses on the version with the best test results: YOLO8-l-seg, a YOLO8 \cite{YOLOv8} variant designed for instance segmentation tasks. We use the implementation provided by Ultralytics\footnote{\url{https://docs.ultralytics.com/models/yolov8}, last accessed 26 May 2026.}.

\textbf{RT-DETR}. In terms of speed and efficiency, transformer-based detectors (DETRs) have lagged behind other detectors, such as those in the YOLO family. This is important not only for real-time applications but also for the detection of text lines in a large archive with many document pages, which benefits significantly from fast results. For this reason, we also tested RT-DETR \cite{RT-DETR}, which, thanks to changes in its architecture, is supposed to keep up with YOLO models. RT-DETR is an object detection network only, hence it produces bounding boxes instead of fine-grained segmentation masks. Again, we are using the "large" sized implementation by Ultralytics\footnote{\url{https://docs.ultralytics.com/models/rtdetr}, last accessed 26 May 2026.}.

\subsection{HTR CRNN Model}
\label{sec:HTR Model Architectures}

State-of-the-art (SOTA) HTR architectures that do not rely on massive pre-training datasets are still predominantly based on CRNN models. The authors in~\cite{Corbille25} introduced a center-loss-based CRNN architecture that achieves SOTA performance. Although the center-loss formulation improves recognition performance, it also increases the training complexity and time. 
In this work, we use the CRNN-based HTR architecture proposed in~\cite{retsinas2022best}, which is lightweight architecturally similar to~\cite{Corbille25} but does not incorporate the center-loss objective. It remains competitive on the IAM dataset while being simpler to train.

The CRNN-based HTR architecture proposed in~\cite{retsinas2022best} begins with a \(7\times7\) convolutional layer with 32 output channels, followed by a sequence of ResNet blocks. Specifically, the network contains two ResNet blocks with 64 output channels, four blocks with 128 channels, and four with 256 channels. Column-wise max pooling is then applied to aggregate the spatial features. The recurrent module consists of three Bi-LSTM layers with a hidden dimension of 256.

\subsection{Text Line Segmentation Training Protocol}
\label{sec: Training Protocol}

First, we train all three methods on the four training datasets and evaluate their segmentation performance in-domain. We test zero-shot out-of-domain on the remaining six datasets to determine their generalization capability.

In the second phase, we evaluate few-shot learning by adding a small number of target-domain samples to improve line detector performance and assess sensitivity to limited target-domain data.

\textbf{Hi-SAM} was trained with 6x A100 40GB GPUs with a batch size of 1 and an initial learning rate (LR) of 1e-4, which was reduced to 1e-5 after a patience of 200 epochs. Following the learning rate reduction, early stopping was applied with an additional patience of 200 epochs. We adopted the original data augmentation strategy used for Hi-SAM, but limited the maximum random rotation to 20$^{ \circ}$. Training Hi-SAM requires document images, polygon annotations for text lines, and corresponding binary masks. Although Hi-SAM is capable of generating the necessary binary masks, it is sensitive to bleed-through text. Therefore, we produce Otsu~\cite{otsu1975threshold}-based binary masks and combine them with the masks produced by Hi-SAM and subsequently remove background regions not covered by the ground-truth line polygons.

Both \textbf{YOLOv8} and \textbf{RT-DETR} were trained for up to 5,000 epochs with early stopping (patience = 250) using the MuSGD optimizer (LR 0.01, momentum 0.9), batch size of 8, and an input size of 1280 × 1280 pixels.

To ensure a fair comparison, Hi-SAM was initialized using SAM pre-trained weights derived from natural-image training, whereas YOLO and RT-DETR were initialized using weights pre-trained on the COCO dataset. As RT-DETR produces only bounding-box predictions, the segmentation masks generated by Hi-SAM and YOLO are converted into bounding boxes to enable a consistent comparison across models. 

We observed that RT-DETR creates an extensive amount of overlapping predictions. To mitigate this issue, we apply additional NMS as a post-processing step to remove redundant detections at inference.

The optimal confidence and NMS thresholds for YOLO and RT-DETR, as well as the optimal score and NMS thresholds for Hi-SAM, were identified through a grid search that we performed on the source-domain validation set.

For few-shot learning, we created two training sub-splits from each target dataset: one containing 10 sample pages and another containing 50 sample pages, where the 10-page split is a subset of the 50-page split. During training, only the selected few-shot samples were used as training data, while the full validation set of the target dataset was used for model selection.

The total number of training epochs for Hi-SAM was limited to 300. Fine-tuning was performed with an initial learning rate of 1e-4, which was reduced to 1e-5 after 250 epochs. All experiments were conducted using a single NVIDIA A100 GPU. Both YOLO and RT-DETR were trained a total of 8,000 epochs without early stopping and we picked the best model based on the validation mAP. The training was conducted with an AdamW optimizer with an initial learning rate of 0.002 and a momentum of 0.9. Both training and fine-tuning used a single A100 GPU.

\subsection{HTR Training Protocol}
\label{sec: CRNN Training Protocol}

For each dataset, we train a CRNN HTR model using text lines obtained from the ground-truth bounding boxes together with their corresponding transcriptions.
The CRNN models are trained using the Adam optimizer with a batch size of 32 and a learning rate of 5e-4. The training is conducted for 1,500 total epochs. The learning rate is reduced to 5e-5 with a patience of 100 and early stopping is applied with a patience of 200 after the learning rate drop.

\section{Results}
\label{sec:results}

\subsection{Quantitative Evaluation of Line Segmentation}

This section presents the quantitative evaluation of the trained line segmentation models. We report performance using mAP, which measures how accurately the predicted bounding boxes align with the ground-truth line annotations.

\begin{table}[!htbp] %[htbp]
% \begin{sidewaystable}[htbp]
\centering
\caption{Cross-dataset evaluation of three line detectors trained on four datasets and tested on nine datasets. Performance is reported as mAP, where higher value is better.}
\label{tab:0-shot_mAP}
\resizebox{\textwidth}{!}{%
\begin{tabular}{l|l|ccccccccc}
\multicolumn{2}{c|}{}
&\multicolumn{9}{c}{\cellcolor{gray!20}\textbf{Tested on}} \\
\hline
\textbf{Training Dataset} 
& \makecell{\textbf{Line}\\\textbf{Detector}}
& \textbf{Bentham} 
& \makecell{\textbf{NorHand}\\\textbf{72702}}
& \makecell{\textbf{NorHand}\\\textbf{331861}}
& \makecell{\textbf{Riksarkivet}\\\textbf{BK 1}}
& \makecell{\textbf{Riksarkivet}\\\textbf{BK 2}}
& \makecell{\textbf{Riksarkivet}\\\textbf{GBG 1}}
& \makecell{\textbf{Riksarkivet}\\\textbf{GBG 2}}
&\makecell{\textbf{READ}\\\textbf{2016}}
& \makecell{\textbf{Nuremberg}\\\textbf{LB 3}} \\
\hline
NorHand-331861 & Hi-SAM &49.64 & 58.00 & \cellcolor{gray!20}75.53 & 68.05 & 62.02 & 55.62 & 47.15 & 28.77 & 41.29  
\\
 & YOLO   &39.51 & 52.89 & \cellcolor{gray!20}81.67 & 74.78 & 71.31 & 58.53 & 48.72 & 21.31 & 19.86  
\\
 & RT-DETR &12.05 & 15.81 & \cellcolor{gray!20}37.22 & 16.58 & 27.50 & 15.85 & 13.10 & 07.41  & 05.27   
\\
\hline
READ 2016 & Hi-SAM & 32.17 & 35.85 & 41.74 & 29.22 & 33.85 & 38.43 & 33.75 & \cellcolor{gray!20}79.01 & 49.08 
\\
 & YOLO    & 25.26 & 13.98 & 13.35 & 24.43 & 21.74 & 34.94 & 27.65 & \cellcolor{gray!20}81.12 & 28.22  
\\
 & RT-DETR &20.88 & 37.64 & 25.59 & 37.02 & 41.12 & 36.37 & 29.83 & \cellcolor{gray!20}67.90 & 32.49  
\\
\hline
Nuremberg LB 3 & Hi-SAM &30.65 & 51.58 & 50.62 & 46.95 & 55.55 & 54.71 & 37.13 & 38.93 & \cellcolor{gray!20}78.25  
\\
 & YOLO   &20.70 & 44.88 & 24.63 & 54.78 & 51.15 & 49.75 & 29.18 & 18.72 & \cellcolor{gray!20}74.78  
\\
 & RT-DETR &14.69 & 16.91 & 01.70  & 08.98  & 04.23  & 29.16 & 15.74 & 10.94 & \cellcolor{gray!20}50.31  
\\
\hline
Mix & Hi-SAM &45.32 & 54.04 & \cellcolor{gray!20}71.58 & 40.64 & 42.16 & 44.18 & 37.08 & \cellcolor{gray!20}75.26 & \cellcolor{gray!20}77.82 
\\
 & YOLO   &30.44 & 48.46 & \cellcolor{gray!20}75.89 & 51.41 & 60.40 & 61.24 & 50.04 & \cellcolor{gray!20}71.87 & \cellcolor{gray!20}76.01  
\\
 & RT-DETR &27.92 & 28.87 & \cellcolor{gray!20}37.94 & 25.56 & 38.10 & 20.04 & 21.31 & \cellcolor{gray!20}44.59 & \cellcolor{gray!20}26.77  \\
\hline
\end{tabular}%
}
\end{table}

We trained the three line segmentation models described in Section~\ref{sec:Line Detector Model Architectures} on the four training datasets: NorHand-331861, READ 2016, Nuremberg LB 3, and the Mix dataset. After that, we evaluated their performance on all nine datasets described in Section~\ref{sec:datasets} and report the results in Table~\ref{tab:0-shot_mAP}. As would be expected, the models generally perform best in the in-domain setting (shaded). 
Hi-SAM and YOLO achieved higher performance than RT-DETR.

For the out-of-domain (OOD) setting (not shaded), we evaluate the generalizability of the line detectors by training the model on one dataset and testing on the remaining eight datasets, excluding the training dataset. For the Mix dataset, we consider only six test datasets. As shown in Table~\ref{tab:0-shot_mAP}, when Hi-SAM is trained on NorHand-331861, READ 2016, or Nuremberg LB 3 and evaluated on out-of-domain datasets, the mAP values are below 50\% in most cases. Only two out of 24 out-of-domain evaluations achieve a mAP above 60\%, both when Hi-SAM is trained on NorHand-331861 and tested on Riksarkivet BK~1~\&~2, suggesting limited but dataset-dependent generalization.

YOLO follows a similar trend, with only the same two cases achieving a mAP above 70\%. Hi-SAM outperforms YOLO in most OOD settings. The only exceptions were five cases: when YOLO was trained on NorHand-331861 and evaluated on Riksarkivet BK~1~\&~2, and Riksarkivet GBG~1~\&~2; and when YOLO was trained on Nuremberg LB 3 and evaluated on Riksarkivet BK 1. For RT-DETR, all OOD cases achieved a mAP below 50\%.

We then trained the models on the Mix dataset to improve domain generalizability. In this setting, only six datasets were considered out-of-domain, since the Mix dataset includes training samples from NorHand-331861, READ 2016, and Nuremberg LB 3. As shown in Table~\ref{tab:0-shot_mAP}, only YOLO achieved a mAP above 60\% when evaluated on the Riksarkivet BK 2 and Riksarkivet GBG 1 datasets.

Overall, these results indicate that all three models do not generalize robustly for text line segmentation across unseen datasets.

\subsection{Qualitative Evaluation of Line Segmentation}

\begin{figure}[t] %[H]
    \centering
    \includegraphics[width=1\linewidth]{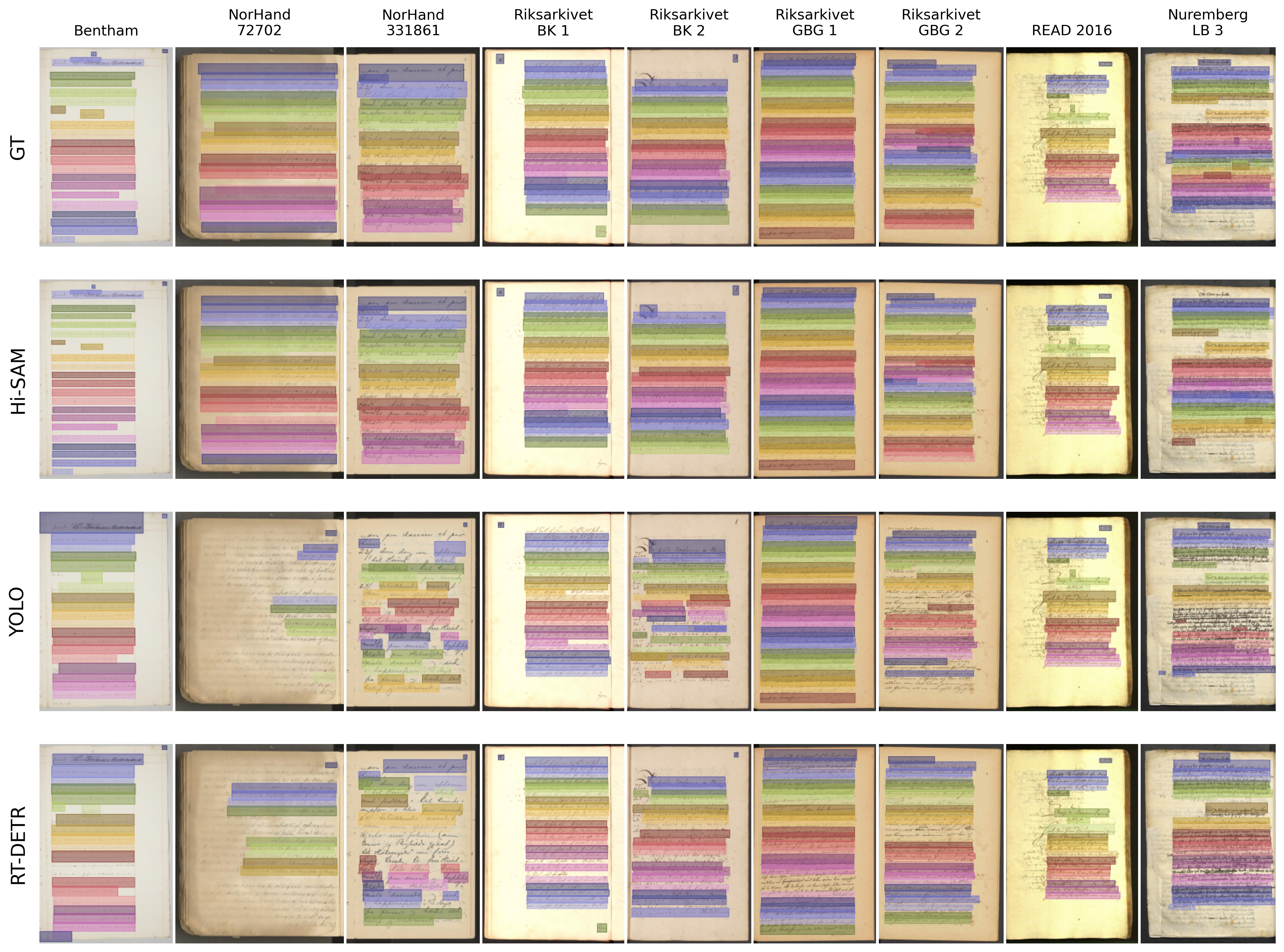}
    \caption{Qualitative comparison. The first row shows ground-truth bounding boxes; the remaining rows show the corresponding predictions from each line detection model on samples from the nine datasets used in this study.}
    \label{fig:qualitative_samples}
\end{figure}

We conducted a qualitative analysis to further validate the quantitative results discussed above. All models in Table~\ref{tab:0-shot_mAP} generally achieved low mAP scores, indicating limited cross-domain generalization. However, the qualitative results in Figure~\ref{fig:qualitative_samples} show that model behavior varies across datasets. In particular, the Hi-SAM predictions, shown in the second row, align more closely with the ground-truth bounding boxes shown in the first row. YOLO and RT-DETR failed to segment text lines accurately in most cases, highlighting that not all line detection models are equally poor in terms of qualitative segmentation performance. 

\subsection{Investigation into mAP as Evaluation Metric}
\label{sec:mAP-CER-metric}

\begin{figure} %[H]
    \centering
    \includegraphics[width=0.7\linewidth]{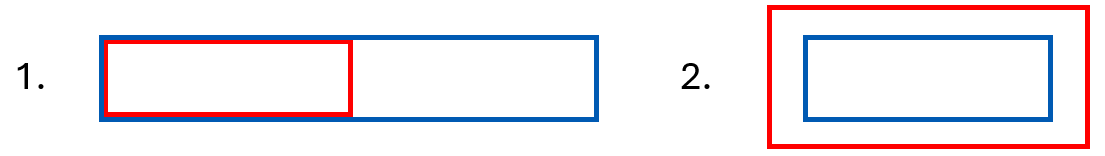}
    \caption{Two ways IoU 0.5 overlap can occur. The blue box is the ground-truth bounding box. The red box is the predicted bounding box.}
    \label{fig:IoU_0.5_overlap}
\end{figure}

To better understand the discrepancy between the qualitative and quantitative results, we examined the mAP calculation procedure. mAP is computed by first measuring the Intersection over Union (IoU) between each predicted bounding box and its corresponding ground-truth box. A set of IoU thresholds is then applied, as described in Section~\ref{sec: Evaluation Metrics}, to classify detections as true positives or false positives. However, for a given IoU threshold, there are two possible outcomes for each predicted bounding box. For example, as illustrated in Figure~\ref{fig:IoU_0.5_overlap}, a prediction that covers only half of the ground-truth region (case 1) has an IoU measure of 0.5, but is clearly insufficient for line level HTR and may result in increased CER. Conversely, a prediction that slightly extends beyond the ground-truth boundaries on all four sides (case 2) can also achieve an IoU of 0.5 while having minimal impact on HTR performance and only marginally affecting the CER as the predicted box covers the full text line. Therefore, even when mAP values are low, the predicted bounding boxes may still be adequate for downstream HTR tasks, highlighting a limitation when relying solely on IoU-based detection metrics.

\subsection{Quantitative Evaluation of Recognition}

\begin{table}[!htbp]
% \begin{sidewaystable}[htbp]
\centering
\caption{Cross-dataset evaluation of three line detectors trained on four datasets and tested on 9 datasets. Performance is reported as CER\%, where lower values indicate better performance.}
\label{tab:0-shot_CER}
\resizebox{\textwidth}{!}{%
\begin{tabular}{l|l|ccccccccc|c}
\multicolumn{2}{c|}{}
& \multicolumn{9}{c|}{\cellcolor{gray!20}\textbf{Tested on}} & \\
\hline
\textbf{Training Dataset} 
& \makecell{\textbf{Line}\\\textbf{Detector}}
& \textbf{Bentham} 
& \makecell{\textbf{NorHand}\\\textbf{72702}}
& \makecell{\textbf{NorHand}\\\textbf{331861}}
& \makecell{\textbf{Riksarkivet}\\\textbf{BK 1}}
& \makecell{\textbf{Riksarkivet}\\\textbf{BK 2}}
& \makecell{\textbf{Riksarkivet}\\\textbf{GBG 1}}
& \makecell{\textbf{Riksarkivet}\\\textbf{GBG 2}} 
& \makecell{\textbf{READ}\\\textbf{2016}}
& \makecell{\textbf{Nuremberg}\\\textbf{LB 3}}
& \makecell{\textbf{OOD AVG}\\\textbf{gap with}\\\textbf{Baseline}} \\
\hline
Baseline & GT-Boxes 
& {\textbf{2.22}} 
& {\textbf{1.61}} 
& {\textbf{3.09}} 
& {\textbf{3.40}} 
& {\textbf{2.48}} 
& {\textbf{4.60}} 
& {\textbf{2.96}} 
& {\textbf{3.99}} 
& {\textbf{5.02}} 
& 
\\
\hline
NorHand-331861 & Hi-SAM 
& 5.03 
& 3.56 
& {\cellcolor{gray!20}3.65} 
& 4.72 
& 3.53 
& 8.81 
& 13.82 
& 13.46 
& 14.05 
& {\textbf{5.09}} 
\\
 & YOLO   
& 7.47 
& 6.64 
& {\cellcolor{gray!20}3.48} 
& 4.93 
& 5.93 
& 9.67 
& 14.13 
& 14.82 
& 39.90 
& 9.65 
\\
 & RT-DETR 
& 26.59 
& 48.81 
& {\cellcolor{gray!20}24.12} 
& 56.77 
& 44.37 
& 63.90 
& 64.67 
& 70.25 
& 86.14 
& 54.40 
\\
\hline
READ 2016 & Hi-SAM 
& 5.34 
& 4.08 
& 6.52 
& 4.74 
& 3.73 
& 8.81 
& 12.39 
& {\cellcolor{gray!20}4.40} 
& 8.35 
& {\textbf{3.57}} 
\\
 & YOLO   
& 8.28 
& 37.71 
& 34.36 
& 9.53 
& 14.47 
& 17.81 
& 25.59 
& {\cellcolor{gray!20}4.31} 
& 37.56 
& 19.99 
\\
 & RT-DETR 
& 9.32 
& 23.90 
& 31.92 
& 8.67 
& 13.79 
& 22.70 
& 26.27 
& {\cellcolor{gray!20}6.90} 
& 30.49 
& 17.71
\\
\hline
Nuremberg LB 3 & Hi-SAM 
& 4.81 
& 4.51 
& 10.15 
& 4.02 
& 3.71 
& 9.17 
& 23.58 
& 6.53 
& {\cellcolor{gray!20}6.91} 
& {\textbf{5.27}} 
\\
& YOLO   
& 15.70 
& 11.50 
& 34.78 
& 5.40 
& 13.02 
& 9.08 
& 34.23 
& 19.42 
& {\cellcolor{gray!20}6.63} 
& 14.85 
\\
& RT-DETR 
& 46.89 
& 37.83 
& 75.44 
& 67.42 
& 75.00 
& 46.51 
& 57.55 
& 58.08 
& {\cellcolor{gray!20}24.85} 
& 55.05 
\\
\hline
Mix & Hi-SAM 
& 4.44 
& 3.36 
& {\cellcolor{gray!20}3.69} 
& 4.20 
& 3.14 
& 7.77 
& 11.76
& {\cellcolor{gray!20}4.56} 
& {\cellcolor{gray!20}6.60} 
& {\textbf{2.90}} 
\\
& YOLO   
& 5.48 
& 10.81 
& {\cellcolor{gray!20}3.96} 
& 4.09 
& 3.67 
& 7.53 
& 11.87 
& {\cellcolor{gray!20}4.22} 
& {\cellcolor{gray!20}7.66} 
& 4.36 
\\
& RT-DETR 
& 20.03 
& 28.28 
& {\cellcolor{gray!20}32.04} 
& 21.36 
& 20.21 
& 48.65 
& 43.22 
& {\cellcolor{gray!20}31.80} 
& {\cellcolor{gray!20}51.87} 
& 27.41 \\
\hline
\end{tabular}%
}
\end{table}

We used the same bounding boxes generated by the three line detectors that were used to calculate the mAP in Table~\ref{tab:0-shot_mAP}, and now recognize the text content using the CRNN HTR model presented in Section~\ref{sec:HTR Model Architectures}. We calculate the CER and present the results in Table~\ref{tab:0-shot_CER}. The top row of Table~\ref{tab:0-shot_CER} presents the baseline CER values when the HTR model is trained with "optimum" text lines from the GT bounding boxes and evaluated on the GT testing split of each dataset.

For the in-domain case, when the models are trained and tested with the test split from the same document collection, the CER results (Table~\ref{tab:0-shot_CER} shaded) are very close to the baseline with text line boxes from Hi-SAM and YOLO giving a maximum gap of 0.56 percentage points (pp). However, RT-DETR CER results are significantly higher when the models are trained with NorHand-331861 and Nuremberg LB 3 indicating it is unsuitable even for the in-domain setting.

The CER values for the out-of-domain evaluation setting, where the line detector models are trained on one dataset and tested on the remaining datasets, are presented in Table~\ref{tab:0-shot_CER} (not shaded). Additionally, the last column reports the average CER gap between the OOD predictions and the GT baseline. When Hi-SAM is trained on NorHand-331861, READ 2016, Nuremberg LB 3, or the Mix dataset and evaluated on out-of-domain datasets, the average CER gaps relative to the baseline are 5.09pp, 3.57pp, 5.27pp and 2.90pp, respectively.
The OOD YOLO average CER gaps are 9.65pp, 19.99pp, 14.85pp and 4.36pp, respectively.
The RT-DETR reported the highest OOD average CER gaps with 54.40pp, 17.71pp, 55.05pp and 27.41pp, thus indicating poor generalizability.

We observed that when YOLO is trained on the Mix dataset, the gap between the baseline CER and the CER obtained from predicted line boxes is substantially reduced. In this setting, the OOD average CER gap is 4.36pp which is now highly competitive with Hi-SAM. This suggests that, compared with Hi-SAM, YOLO benefits more from diverse and generalizable training data.

\subsection{Impact of Few-shot Learning}

We observed the strongest generalizability when the models were trained on the Mix dataset, especially Hi-SAM and YOLO. In this section, we fine-tune them using 10 and 50 target-domain few-shot sample pages to evaluate whether limited in-domain supervision can further improve line detector performance. Given its comparatively poor performance across the zero-shot evaluations, RT-DETR was excluded from the few-shot fine-tuning stage.

\begin{table}[!htbp]
\centering
\caption{Few-shot mAP\% evaluation of Hi-SAM and YOLO line detectors pre-trained on the Mix dataset and fine-tuned on six target datasets. Zero-shot rows report the corresponding results from Table~\ref{tab:0-shot_mAP}. Higher values indicate better performance.}
\label{tab:mix_few_shot_mAP}
\resizebox{\textwidth}{!}{%
\begin{tabular}{l|l|cccccc}
\multicolumn{2}{c|}{}
& \multicolumn{6}{c}{\cellcolor{gray!20}\textbf{Tested on}} \\
\hline
\textbf{} 
& \makecell{\textbf{Line}\\\textbf{Detector}}
& \textbf{Bentham}
& \makecell{\textbf{NorHand}\\\textbf{72702}}
& \makecell{\textbf{Riksarkivet}\\\textbf{BK}}
& \makecell{\textbf{Riksarkivet}\\\textbf{BK 2}}
& \makecell{\textbf{Riksarkivet}\\\textbf{GBG 1}}
& \makecell{\textbf{Riksarkivet}\\\textbf{GBG 2}} \\
\hline
\rowcolor{gray!30}
Zero-shot & Hi-SAM 
& 45.32 
& 54.04 
& 40.64 & 42.16 & 44.18 & 37.08  
\\
Few-shot 10 & Hi-SAM 
& 64.32 
& 54.64  
& 73.91 & 71.75 & 62.77 & 53.99  
\\
Few-shot 50 & Hi-SAM 
& 68.10 
& 60.64 
& 75.44 & 72.81 & 65.50 & 59.12  
\\
\hline
\rowcolor{gray!30}
Zero-shot & YOLO 
& 30.44 
& 48.46 
& 51.41 & 60.40 & 61.24 & 50.04  
\\
Few-shot 10 & YOLO 
& 62.04 
& 57.25  
& 82.14 & 80.49 & 67.86 & 61.02 
\\
Few-shot 50 & YOLO 
& 69.05 & 62.23 & 83.73 & 83.86 & 71.64 & 65.41 \\
\hline
\end{tabular}%
}
\end{table}

As shown in Table~\ref{tab:mix_few_shot_mAP}, both Hi-SAM and YOLO achieve substantial mAP improvements when fine-tuned with even just 10 few-shot samples from the out-of-domain target datasets. Increasing the few-shot set from 10 to 50 samples provides additional gains, but they appear smaller considering the additional 40 pages of zoning an archive would have to complete on the target domain. This suggests the sufficiency of 10 few-shot samples.

\begin{table}[!htbp]
\centering
\caption{Few-shot CER\% evaluation of line detectors pre-trained on the Mix dataset and fine-tuned on six OOD datasets. Zero-shot rows report the corresponding results from Table~\ref{tab:0-shot_CER}. Lower values indicate better performance. The GT row reports the performance of the HTR model when trained using ground-truth line bounding boxes.}
\label{tab:mix_few_shot_CER}
\resizebox{\textwidth}{!}{%
\begin{tabular}{l|l|cccccc|c}
\multicolumn{2}{c|}{}
& \multicolumn{6}{c}{\cellcolor{gray!20}\textbf{Tested on}} &  \\
\hline
\textbf{} 
& \makecell{\textbf{Line}\\\textbf{Detector}}
& \textbf{Bentham}
& \makecell{\textbf{NorHand}\\\textbf{72702}}
& \makecell{\textbf{Riksarkivet}\\\textbf{BK 1}}
& \makecell{\textbf{Riksarkivet}\\\textbf{BK 2}}
& \makecell{\textbf{Riksarkivet}\\\textbf{GBG 1}}
& \makecell{\textbf{Riksarkivet}\\\textbf{GBG 2}} 
& \makecell{\textbf{OOD AVG}\\\textbf{gap with}\\\textbf{Baseline}} \\
\hline
Baseline & GT-Boxes 
& \textbf{2.22} 
& \textbf{1.61} 
& \textbf{3.40} 
& \textbf{2.48} 
& \textbf{4.60} 
& \textbf{2.96} 
& 
\\
\hline
\rowcolor{gray!30}
Zero-shot & Hi-SAM & 4.44 & 3.36 & 4.20 & 3.14 & 7.77 & 11.76 & 2.90
\\
Few-shot 10 & Hi-SAM & 3.31& 2.72   & 3.89 & 2.64 & 6.59 & 8.25   & 1.69 
\\
Few-shot 50 & Hi-SAM & 3.22 & 2.56   & 3.57 & 2.69 & 6.50 & 5.89  & 1.19 
\\
\hline
\rowcolor{gray!30}
Zero-shot & YOLO & 5.48 & 10.81 & 4.09 & 3.67 & 7.53 & 11.87   & 4.36
\\
Few-shot 10 & YOLO & 3.79 & 3.08   & 3.80 & 2.93 & 7.12 & 8.80  & 2.04 
\\
Few-shot 50 & YOLO & 3.64 & 2.79  & 3.49 & 2.48 & 7.07 & 6.32   & 1.42 \\
\hline
 COCO + Few-shot 10& YOLO& 6.12& 94.56& 3.75& 16.03& 8.27& 9.93& 20.23\\
 COCO + Few-shot 50& YOLO& 4.31& 2.65& 3.55& 2.55& 7.13& 5.60& 1.42\\
 \hline
\end{tabular}%
}
\end{table}

Fine-tuning on the target datasets reduces the CER gap relative to the baseline, as shown in Table~\ref{tab:mix_few_shot_CER}. For Hi-SAM, the average gap decreases from 2.90pp to 1.69pp with target-domain fine-tuning with 10 samples. Increasing the few-shot set to 50 samples provides only a marginal improvement, reducing the average gap to 1.19pp. 
YOLO, on the other hand, is more sensitive to few-shot fine-tuning. In the most extreme case, when fine-tuned on NorHand-72702, the CER decreases from 10.81\% to 3.08\% using only 10 few-shot samples. On average, using 10 few-shot samples reduces the CER gap between the baseline and prediction-based results to 2.04pp. Increasing the number of samples to 50 further reduces the gap to 1.42pp, showing a similar diminishing-return pattern to that observed for Hi-SAM.

\section{Discussion}
\label{sec:discussion}

We observed that Hi-SAM generalizes well across the out-of-domain target datasets, regardless of the dataset used for training. In contrast, YOLO shows limited generalization when trained on a single dataset. However, training YOLO on the Mix dataset substantially improves its cross-domain performance by introducing greater domain diversity during training. When the Mix-trained YOLO model is further fine-tuned with only 10 target-domain samples, its performance becomes marginally closer to that of Hi-SAM.

\subsection{Generalization Capabilities}

In terms of pure generalization, \textit{i.e.} training on one dataset and then making predictions on various others, the Hi-SAM model demonstrates remarkable ability. As illustrated qualitatively in Figure \ref{fig:qualitative_samples}, Hi-SAM reliably recognizes complete lines, even in visually different domains. Although the predicted text line regions are somewhat more extensive than the corresponding ground-truth on which our HTR models were trained, Hi-SAM's recognition results are significantly better.

On the other hand, both YOLO and RT-DETR exhibit noticeable difficulties with direct zero-shot generalization. For example, YOLO has difficulty capturing complete lines of text, so only sections are often detected. This occurs despite the use of appropriate data augmentation, which presents the network with lines of text spanning the entire width of the input image. Variations in the structure and color of the substrate also hinder recognition. RT-DETR typically recognizes the full line of text but often skips entire lines and paragraphs.

\subsection{mAP vs CER as Evaluation Metric}

These results further highlight the importance of using an appropriate metric to evaluate prediction quality. As discussed in more detail in Section \ref{sec:mAP-CER-metric}, it is crucial for downstream text recognition that the text is captured in its entirety, even if the bounding box noticeably exceeds the ground-truth box. Hi-SAM's results demonstrate this. Its mAP scores indicate segmentation results comparable to those of a YOLO model. However, a qualitative assessment and subsequent CER analysis show that Hi-SAM's boxes are significantly better suited for text recognition because they capture the full text. YOLO and RT-DETR fail to recognize substantial portions of the document.

\subsection{Fine-tuning Impact}

Our primary interest in fine-tuning was focusing on whether the YOLO model, which is significantly smaller and thus faster, could catch up to Hi-SAM's performance with a relatively small amount of additional training data, despite lagging behind Hi-SAM in terms of zero-shot generalization.

The results suggest that it can. Even with rather low-cost annotation of just ten pages from the desired target dataset, results can match the zero-shot performance of Hi-SAM. Specifically, as shown in Table \ref{tab:mix_few_shot_CER}, the average CER deviation from the ground-truth baseline is reduced from 4.36pp to 2.04pp with just ten annotated pages and to 1.42pp with 50 pages. Hi-SAM continues to achieve better results with 10 (1.69pp) and 50 (1.19pp) pages of fine-tuning but does so at significantly higher training costs. The qualitative improvement over the zero-shot generalization result is also less pronounced.

For a direct comparison, we also initialized a YOLO model with standard COCO pre-training and trained it on the 10- and 50-shot pages, presented in the last two rows of Table \ref{tab:mix_few_shot_CER}. With only ten pages, YOLO's results are noticeably unstable and fail to produce usable results for some datasets. However, when there is sufficient training data available, competitive performance can even be achieved without pre-training on other historical documents.

\subsection{Resource Requirements}

Apart from segmentation performance, size is one of the most significant differences between the Hi-SAM and YOLO models. Hi-SAM is far more complex in memory size and number of parameters and thus challenging to train. Nevertheless, Hi-SAM delivers a greater generalization capability than the other models.

\begin{table}[!htbp]
    \centering
    \caption{Overview of run times of the tested models in different use case environments, investigating the loading time for models as well as training and inference performance.}
    \resizebox{\textwidth}{!}{%
    \begin{tabular}{ll||c|c|c||c|c|c}
    \multicolumn{2}{c||}{} 
    & \multicolumn{3}{c||}{\cellcolor{gray!20}\textbf{Training}} 
    & \multicolumn{3}{c}{\cellcolor{gray!20}\textbf{Inference}} \\
         &  &  GPU Memory&  \# Param. & \makecell{Training\\in (GPU) hours} & GPU Memory &  Model Load Time  & \makecell{Inference per Page\\Batch Size = 1}\\ 
         \hline
         \hline
         % Berzelius (
         A100 &  Hi-SAM & 40 GB &  \makecell{699.2 M\\(62.2 trainable)} & \makecell{690 GPUh (with SAM weights)\\380 GPUh (with Hi-SAM weights)} & 13 GB & 9.75 s & 1.6 s\\
         &  YOLO& 26 GB&  46.0 M & 2.5 GPUh (avg 3100 epochs) &  1.5 GB& 0.5 s & 0.12 s\\
         &  RT-DETR&  23 GB&  32.8 M & 2.0 GPUh (avg 2500 epochs) & 1.2 GB&  0.5 s & 0.13 s\\    \hline
         
         RTX 4080&  Hi-SAM & -&  \makecell{699.2 M\\(62.2 trainable)} & -& - &  4.97 s& 1.5 s\\ 
         &  YOLO &  -&  46.0 M &  50 h (for 3100 epochs) & -&  0.10 s& 0.11 s\\
         &  RT-DETR&  -&  32.8 M &  - & -&  0.10 s& 0.11 s\\ \hline
         
         Laptop (CPU)&  Hi-SAM & - &  \makecell{699.2 M\\(62.2 trainable)} & -& - & 0.16 s & 100.13 s \\
         &  YOLO &  - & 46.0 M &  7500 h (est. extrapolation for 3100 epochs) & -&  0.15 s& 4.75 s\\
         & RT-DETR& - & 32.8 M & 3800 h (est. extrapolation for 2500 epochs) & -& 0.17 s&4.66 s\\ \hline
    \end{tabular}
    }
    \label{tab:resources}
\end{table}

Table \ref{tab:resources} compares training and inference performance of three different computational scenarios: a high-performance cluster designed for machine learning tasks with A100 GPU nodes available, a desktop computer with a dedicated graphics card (AMD Ryzen 9 7950X3D @ 4.20~GHz, GeForce RTX 4080, 64~GB DDR5-RAM), and a laptop without a dedicated graphics card (Intel Core Ultra 7 155 U @ 4.8~GHz, 64~GB DDR5-RAM). In all cases we trained the models on the 350 pages of the READ training set, and we computed the inference on the 50 pages in the READ test set before dividing by the number of pages.

Hi-SAM requires substantially more computational resources than the other models. Initializing Hi-SAM with pre-trained weights reduces the training cost by nearly half while maintaining similar generalization performance (see Supplementary Material). In resource-constrained settings, particularly with limited GPU availability, YOLO provides a competitive alternative, achieving strong results when trained on the Mix dataset and fine-tuned with only 10 target-domain pages.

As expected, a dedicated graphics card greatly impacts training and inference times. Therefore, purchasing an older graphics card can be useful when dealing with large document collections. However, it is unclear whether the time investment required to train a specialized YOLO model is always worthwhile. Given Hi-SAM's strong generalization results, testing the pretrained Hi-SAM model on your dataset before investing more time in creating training data and fine-tuning seems like a reasonable alternative.

\section{Conclusion}
\label{sec:conclusion}

Our experimental study shows that text line segmentation does not require extensive collection of new training data for each new dataset.

The complex yet highly capable Hi-SAM model, in particular, has demonstrated that it can reliably localize text lines across a wide range of potential target domains with highly varied visual layouts, even when trained on a single initial dataset. This makes Hi-SAM a promising zero-shot foundation model for previously untapped datasets, enabling the direct extraction of text lines from document pages.

Our fine-tuning experiments have also shown that simpler, smaller models, such as YOLO, can be trained to specialize in a specific dataset using affordable hardware architectures and achieve competitive results with only a small amount of annotated data. As few as ten pages are sufficient to fine-tune a model to the characteristics of a given dataset.

The experiments conducted here have primarily focused on documents with a rather simple layout: single-column text arranged in neat horizontal lines with minimal marginalia. Detecting and recognizing more complex layouts, such as those with curved or rotated text lines or tabular structures, increases the complexity of the problem. This issue will be examined in greater detail in future research.

In this study we focused on the generalization capabilities of text line segmentation models. However, these models constitute only the first step in an HTR pipeline. Therefore, further research to investigate and develop domain adaptation and generalization for HTR models to determine and reduce the effort required for data acquisition in this context is also needed.

\section*{Acknowledgements}

This work was partially supported by the Wallenberg AI, Autonomous Systems and Software Program (WASP), funded by the Knut and Alice Wallenberg Foundation, with counterpart funding from Luleå University of Technology (LTU).
It was also supported by Kempestiftelserna grant CSMK23-0109. Computations were enabled by the Berzelius resource at the Swedish National Supercomputer Center.
\newpage
\bibliographystyle{unsrt}
\bibliography{bibliography.bib}  %%% Uncomment this line and comment out the ``thebibliography'' section below to use the external .bib file (using bibtex) .





% \end{thebibliography}

\end{document}